\documentclass{withoutJair} %[screen, review]

\usepackage[T1]{fontenc}
\usepackage{mathpazo}
\usepackage{geometry}
\usepackage{enumerate}
\usepackage{graphicx}
\usepackage{tikz-dependency}
\definecolor{darkblue}{rgb}{0, 0, 0.5}
\definecolor{Highlight}{rgb}{0, 0.6, 0.7}
\usepackage{natbib}
\usepackage{usebib}
\usepackage{subcaption}
\usepackage{pdflscape}
\usepackage[nameinlink]{cleveref}
\usepackage{booktabs}
\usepackage{parskip}
\usepackage{array}
\usepackage{colortbl}

\newcommand{\mt}[1]{\texttt{#1}}

\title{Logic Satisfiability as a Probe for Inductive Biases in Neural Networks}
\title{Propositional Logic for Probing Generalization in Neural Networks}
\setcopyright{cc}
\copyrightyear{2025}
\acmYear{2025}
\acmDOI{10.1613/jair.1.xxxxx}

\author{Anna Langedijk}
\orcid{0009-0000-2479-486X}
\email{annalangedijk@gmail.com}
\affiliation{%
  \institution{University of Amsterdam}
  \city{Amsterdam}
  \country{The Netherlands}
}
\author{Jaap Jumelet}
\orcid{0000-0003-3586-1620}
\email{j.w.d.jumelet@rug.nl}
\affiliation{%
  \institution{CLCG, Rijksuniversiteit Groningen}
  \city{Groningen}
  \country{The Netherlands}
}
\author{Willem Zuidema}
\orcid{0000-0002-2362-5447}
\email{w.h.zuidema@uva.nl}
\affiliation{%
  \institution{ILLC, University of Amsterdam}
  \city{Amsterdam}
  \country{The Netherlands}
}

\begin{document}
\begin{abstract}
The extent to which neural networks are able to acquire and represent symbolic rules remains  a key topic of research and debate. 
% Much current work focuses on the impressive capabilities as well as often ill-understood failures of full-scale large language models on a wide range of reasoning tasks. 
Much current work focuses on the impressive capabilities of large language models, as well as their often ill-understood failures on a wide range of reasoning tasks. 
In this paper, in contrast, we investigate the generalization behavior of three key neural architectures -- Transformers, Graph Convolution Networks and LSTMs -- in a controlled task rooted in propositional logic. 
The task requires models to generate satisfying assignments for logical formulas, making it a structured and interpretable setting for studying compositionality.
We introduce a balanced extension of an existing dataset to eliminate superficial patterns and enable testing on unseen operator combinations. Using this dataset, we evaluate the ability of the three architectures to generalize beyond the training distribution. While all models perform well in-distribution, we find that generalization to unseen patterns, particularly those involving negation, remains a significant challenge. Transformers fail to apply negation compositionally, unless structural biases are introduced. Our findings highlight persistent limitations in the ability of standard architectures to learn systematic representations of logical operators, suggesting the need for stronger inductive biases to support robust rule-based reasoning.

\end{abstract}

\received{31 May 2025}

% \begin{abstract}
%     This is the abstract
% \end{abstract}

\maketitle

\section{Introduction}

%\section{Introduction}
%\label{sec:intro}

The relation between neural networks and symbolic logic has been a topic of debate for decades, % within Artificial Intelligence and Cognitive Science.
centered around the question whether sub-symbolic systems are able to build up robust representations of symbolic rules by learning from data.
% The central questions in that debate have been variants of the yes/no questions 'Can neural networks represent logical rules?' and 'Can they learn them through backpropagation?', with many different neural networks architectures and logics considered \cite{}. 
%
With the recent advances in deep learning, and specifically the successes of `Large Reasoning Models' such as OpenAI-O3 and DeepSeek-R1 \cite{deepseek-ai_deepseekr1incentivizingreasoning_2025}, the debate is shifting. 
When reasoning capabilities of state-of-the-art language models are rigorously evaluated, researchers report both impressive generalization beyond the training data and unexpected failure models, where the models seem to rely on heuristics and spurious correlations instead of learning generalizable rules \citep{mccoy_rightwrongreasons_2019,hossain_analysisnaturallanguage_2020,helwe_reasoningtransformerbasedmodels_2021,mondorf_accuracyevaluatingreasoning_2024,wan_triggeringlogical_2024}.
%As it gets more widely accepted that these models to generalize beyond their training data and do more than interpolation between observed reasoning patterns, 
These unexpected failures highlight the fact that, at a fundamental level, the reasoning abilities of large neural models remain insufficiently understood.

While numerous efforts have been made to study reasoning abilities in the context of LLMs \citep{zhang_paradoxlearningreason_2022, parmar-etal-2024-logicbench, liu2025synlogicsynthesizingverifiablereasoning, hazra_havelargelanguage_2025}, we suspect that a rigorous understanding of reasoning and generalization abilities can only be gained in a fully controlled setting.
Similar questions about the intrinsic generalization capacity of neural models have been addressed by testing them on simple synthetic tasks, where one has full control over the training data \citep{hupkes_visualisationdiagnosticclassifiers_2018, hao_formallanguagerecognition_2022, jumelet_transparencysourceevaluating_2023}. 
Evaluating the behaviour of the model on structurally different (out-of-distribution) test sets can then lead to insights about its systematic capabilities.
Furthermore, evaluating model behaviour in such a controlled setting also allows us to investigate the \textit{inductive biases} of different model architectures \citep{DBLP:journals/corr/abs-2006-00555, tay-etal-2023-scaling}.

In this paper, we aim to contribute a small step in this research program, with an in-depth case study that uses synthetic training and testing data from the domain of propositional logic. 
This domain is very restricted, but fundamental within symbolic logic and useful in practice through SAT solving, as well as helpful in revealing the methodological challenges when combining logic and deep learning.
More specifically, we study models trained on a sequence-to-sequence task, in which models must generate a possible world that satisfies an input formula in propositional logic. 
% For instance, $a=true, b=false$ is a possible output for the sentence $\neg (b \wedge a)$ \cite{hahn_teachingtemporallogics_2021}.
Propositional logic is a well-defined task that requires systematic understanding: propositional variables and operators must be recursively combined to form the meaning of the logical formula. 
Previous research shows that Transformer models generalize to out-of-distribution data with varying rates of success \citep[e.g.][]{richardson_pushinglimitsrule_2021,zhang_paradoxlearningreason_2022}.

As a starting point to our study, we analyze and extend the dataset of propositional logic created by \citet{hahn_teachingtemporallogics_2021}.
We identify various shortcomings in this dataset and, balance it to avoid learning shortcuts based on the direction of branching.
This yields a partitioning of the dataset that allows for systematic investigating of the generalization behavior to unseen categories of formulas. 
We then use this dataset to analyze the solutions learned by three popular generative deep learning architectures: a Transformer encoder-decoder (with and without tree-position embeddings), a Graph Convolutional Network and a Long-Short Term Memory Network, and compare their performance.

Our results show that all architectures can learn to generate highly accurate predictions of variable assignments that satisfy a given logic formula: all setups score $\geq$87\% correct on the standard test set. 
However, we also show that such results hide great variation between the three architectures in their abilities to generalize not just to novel formulas, but to novel \emph{categories} of formulas.
More specifically, we examine whether the models can systematically apply their knowledge of \emph{seen} logical operators to \emph{unseen combinations} of operators. 
% By rewriting the data instead of removing it, we can keep the output distribution the same to give the models better chances of succeeding.
We rewrite the data in such a way that certain patterns no longer occur, while the output distribution of variable distributions remains the same, which gives the models better chances of succeeding. 
For example, some models never see the {\sc xor}-operator applied to the {\sc not}-operator, and some models never see the {\sc not}-operator applied to the {\sc or}-operator. We evaluate their performance on such unseen patterns. 

We find that all models trained on limited training sets can in most cases generalize to unseen patterns, suggesting that their learned logical representations are at least partially systematic. %\todo{This also holds for other baseline architectures}
However, when the unseen combination consists of a negation ({\sc not}-operator) applied to another operator, vanilla Transformers fail to generalize to even the simplest sentences containing this combination, demonstrating that specialized negation operators have to be learned for every logical operator. 
% From this we conclude that Transformers are unable to compose negation with an operator
Generalization of the negation operator improves in 2 out of 3 cases when explicit structure is added to the model architecture, in the form of tree-based encodings, recurrence, or graphs. 
However, unlike for other patterns, their performance on the unseen patterns even in simple sentences still lags behind significantly.
These findings cast doubt on whether Transformer-based models can learn systematic representations of negation, a crucial logical operator, from data alone. %\todo{And other architectures.}

\section{Related Work}
%\todo{General related work?}

\paragraph{Learning logic using neural networks}
%\todo{This subsection has to be shorter}

Using neural networks to solve exact problems differs fundamentally from using classical solvers.
Neural networks often require large amounts of training data to learn a task, and even then, their outputs are not guaranteed to be correct.
Therefore, a practical way to use neural methods for logic is to use them in combination with symbolic solvers. For instance, deep neural networks can be used as a variable selection heuristic \citep[e.g.][]{yolcu_learninglocalsearch_}.
Neurosymbolic methods combine reasoning with learning \citep{garcez_neuralsymboliclearningreasoning_2022},  leveraging the ability of neural networks to find complex statistical patterns, without incorporating their noisy behaviour into the output.
In the current paper, however, we focus on attempts to train {\it end-to-end} neural networks to solve logical problems. An advantage of this approach is that, once the network is trained, it can be used to generate possible solutions in polynomial time.

One approach to teaching logic reliably to neural models is to impose the right kind of inductive bias: for example, NeuroSAT \citep{selsam_learningsatsolver_2019} uses a Graph Neural Network (GNN) to solve satisfiability of non-trivial propositional formulas.
Formulas that are limited to a common format (CNF) can be represented as graphs in the network. Relevant clauses and literals are connected by edges in the input graph to the network.
The GNN is trained to only detect whether the input formula is satisfiable or not, which it can do with 85\% accuracy.
For the majority of satisfiable formulas, an actual assignment to the variables can be extracted from the network's hidden activations, even though this was not part of the training objective.
Moreover, the learned weights generalize to formulas that are larger than the network was trained on. This implies that with the right induced structure, neural networks can learn generalizable logical operations to some extent.
%Similarly, \citet{amizadeh_learningsolvecircuitsat_2019} also leverage the inductive bias of GNNs to directly obtain satisfiable assignments from their end-to-end neural SAT solver.

\citet{evans_canneuralnetworks_2018} propose a method called PossibleWorldNet that is not limited to inputs in CNF, but instead uses a custom tree-based recurrent network to solve the task of propositional logical entailment.
PossibleWorldNet also generalizes to sentences of unseen lengths and variable amounts.
%Besides aiding the model with syntax (the tree architecture), they also allow the model to do multiple forward passes to check entailment, all with different random input vectors that represent ``possible worlds'', which significantly improves performance on the classification task. 
Neither NeuroSAT nor PossibleWorldNet explicitly provide the semantics of logic to their machine learning model, but instead make architectural changes so that the neural model may more easily learn the semantics by itself.

Our goals are different from these papers: instead of using neural methods to ultimately improve performance on logical tasks, we use a relatively simple dataset of propositional logic to set out to investigate how and how well neural models can learn systematic representations.
This brings us to the topics of \textbf{compositionality} and {\bf generalization}: do neural models perform well on out-of-distribution data that is longer, more difficult, or structurally different?

\paragraph{Testing for compositional generalization in neural models}
%Describe previous work on systematic generalization testing with artificial datasets.
%To study 
%One theme 
Systematic generalization is a vital aspect of human cognition \citep{FODOR19883,MARCUS1998243,lake_buildingmachinesthat_2016}.
Generalization has become an important way to assess the robustness of reasoning patterns \citep{welleck_symbolicbrittlenesssequence_2022}, and as a general tool to understand the nature of model behaviour \citep{hupkes2022taxonomy}.
% To truly understand a symbolic task such as logic, neural models have to generalize systematically to out-of-distribution data, which is something neural models often struggle to do \citep[e.g.][]{welleck_symbolicbrittlenesssequence_2022}.
One way of testing whether models learn to reason is by simply studying their behaviour on different distributions of data.
Synthetic data allows full control over the train and test distributions, and often comes with the benefit of being fully interpretable, in contrast to objectives like language modeling or complex image classification tasks.
Existing models are evaluated on challenging test sets or datapoints with minimal, systematic edits.
The change in behaviour can then be used to assess how the model interprets these new inputs. If the model behaviour is correct, regardless of its impaired training, the model must have learned some generalizable, compositional information \citep{hupkes_compositionalitydecomposedhow_2020}.

Various tasks to investigate compositional generalization have been proposed in recent years.
\citet{lake_generalizationsystematicitycompositional_2018} introduce a simple synthetic sequence-to-sequence task consisting of commands and actions (for example, the input ``\mt{jump left twice after turn left}'' leads to output ``\mt{LTURN JUMP JUMP}'') in which, during training, certain combinations of actions are never seen. They then show that RNNs fail to compose these unseen functions at test time.
Similarly, \citet{hupkes_compositionalitydecomposedhow_2020} design a string editing task (for example, the input ``\mt{repeat swap A B}'' leads to output ``\mt{B A B A}''), along with several different training splits, to highlight that different kinds of neural architectures generalize and fail in different ways. 
\citet{kim-linzen-2020-cogs} also create a sequence-to-sequence semantic parsing task, COGS, in which the goal is to transform a sentence in English to a corresponding lambda-expression, again leaving systematic gaps in the training data. Performance of RNN-based models as well as Transformers drops drastically when the models are presented with novel examples requiring structural generalization.

% \todo{Within the Natural Language Processing community, .... types of generalization \citep{hupkes2022taxonomy}}

%\todo{Longer outputs = more difficult (Mathias talk), but the splits in COGS where the model has to recombine functions.}

% However, there are several solutions \citep{jiang-bansal-2021-inducing,csordas_devildetailsimple_2022} 

% \todo{A little paragraph on the latest LLMs and generalization testing?}

In this paper, we introduce a synthetic dataset grounded in propositional logic. 
Our experimental setup departs from prior work on compositional generalization in several important ways.
Unlike SCAN-like tasks, our benchmark is more deeply rooted in a well-established academic domain, propositional logic and SAT solving.
Moreover, in contrast to arithmetic-based tasks, propositional variable assignment is inherently \textit{non-local}, presenting a significantly greater challenge for learning systems.
Finally, unlike COGS, our generalization experiments are designed such that task difficulty remains consistent across all training conditions, enabling us to more precisely isolate compositional behavior.

% Note the differences:\todo{Synthesize these bullet items into a paragraph?}

% \begin{itemize}
% \item Our task is non trivial/non arbitrary (unlike e.g. SCAN-like tasks)
% \item Our task is non local (unlike e.g. evaluating arithmetic expressions)
% \item Keep output distribution/difficulty the same (unlike COGS et al)
% \end{itemize}

%Many AI models used today are trained on flat input strings and contain no explicit reasoning modules or logical structure. Thus, they have to solely rely on their unstructured training data to learn any type of logic and reasoning. Models based on the popular Transformer architecture \citep{Vaswani2017}, such as T5, are able to achieve state-of-the-art performance on many downstream NLP tasks where some understanding of logic is required, such as Natural Language Understanding and Question Answering \citep{raffel_exploringlimitstransfer_2020}.
%However, simply scaling up language models does not necessarily improve their reasoning capabilities \citep{kassner_arepretrainedlanguage_2020,yao_selfattentionnetworkscan_2021,rae_scalinglanguagemodels_2022}.

%%% Local Variables:
%%% mode: latex
%%% TeX-master: "main"
%%% End:

\section{Data}\label{sec:replication_methods}

We start our experiments by reproducing the results of \citet{hahn_teachingtemporallogics_2021}, showing that Transformer models, Graph Convolutional Networks and LSTMs can succesfully learn to generate solutions to propositional logic formulae(\cref{sec:replication_methods,sec:experimental_setup}). After our reproduction, we move to the generalization experiments using alternative training sets (\cref{sec:generalization_methods,sec:generalization_results}). 
%
%
%\subsection{Dataset}
For all our experiments, we use a dataset that is based on the \mt{PropRandom35}-dataset generated by Hahn et al. \citet{hahn_teachingtemporallogics_2021}. In this section we briefly describe how the original dataset was generated, and then detail some of its limitations as well as a few small modifications.

\subsection{Original dataset}

The original dataset contains one million randomly generated satisfiable propositional formulas along with a (partial) assignment of variables that satisfies the formula. The task of the model is to generate a satisfying assignment given a propositional formula. 

The target assignments are generated by the \mt{pyaiger} \citep{marcell_vazquez_chanlatte_2018_1405781} library, which is based on Glucose, a modern symbolic SAT solver \citep{audemard_glucosesatsolver_2018}. These assignments can be either partial or complete: a complete assignment assigns a value to every variable in the sentence. A partial assignment only assigns a value to some variables, while still satisfying the formula. For instance, $\{a=true\}$ is a satisfying partial assignment to the sentence $a \vee b$: this sentence will evaluate to true regardless of whether $b$ is set to true or false. (Note that during the training of our deep learning models later on, only one target assignment is available. However, at inference time, we consider any model output that is a satisfying (partial) world for the formula correct, as there are multiple possible outputs for most formulas). 

To avoid the need for parenthesis tokens, input formulas are given in Polish notation, also known as prefix notation: every operator precedes its arguments. This way, there is no ambiguity even without the use of parentheses.
Datapoint examples can be seen in \autoref{tab:data_examples}.

\begin{table}[h]
  \centering\sf
  \begin{tabular}[h]{l @{\hspace{1cm}}l @{\hspace{1cm}}l}
    \toprule
    Propositional formula & Input in Polish notation & Output \\\midrule
    $ \neg a \wedge ( b \vee c )$ & \texttt{\& !\ a | b c}& \texttt{a 0 b 1}\\
    $ a \oplus \neg e $ & \texttt{xor a !\ e} & \texttt{a 1 e 1}\\
    \bottomrule
  \end{tabular}
  \caption{Example datapoints. Outputs are always alphabetically sorted. Note that the first assignment is {\it partial}: the value of \mt{c} may be either \mt{0} or \mt{1}, and may therefore be omitted.}\label{tab:data_examples}
\end{table}

Formulas have a minimum length of 5 and a maximum length of 35. The data is balanced with respect to length: each length $\geq 8$ accounts for roughly 3\% of the formulas in the dataset. 

The formulas contain at most five propositional variables (\texttt{a, b, c, d, e}) and are constructed using five propositional operators: {\sc not} ($\neg$/\texttt{!}), {\sc and} ($\wedge$/\texttt{\&}), {\sc or} ($\vee$/\texttt{|}), {\sc iff} ($\leftrightarrow$/\texttt{<->}) and {\sc xor} ($\oplus$/\texttt{xor}).
Note that all binary operators are commutative, since operators such as implication ($\rightarrow$) are not included, meaning $A \phi B \equiv B \phi A$ holds for all operators $\phi$.

\subsection{Shortcomings and modifications of the original datatset}\label{sec:dataset_issues}

We have studied the dataset made available by Hahn et al., and have found that it is a useful resource for our purposes, although the dataset also has a number of limitations that are useful to state explicitly:
\paragraph{Imbalance} Firstly, we find that the generated data consists of trees that are imbalanced in one direction: right subtrees of the formulas are significantly larger than the subtrees on the right.
This leads to models favouring those imbalanced sentences, and signficantly changes the pattern of results: we find that training on imbalanced data results in a $\geq$25\% drop in accuracy when models are tested on inverted trees. For our experiments, we therefore decide to re-balance all trees, which can be done computationally efficiently by rotating subtrees of existing trees in the dataset. The process and results are described in  \cref{subsec:imbalance_dataset}. 

\paragraph{Absence of double negations} Secondly, we observe that there are no double negations in the dataset: \mt{!\ !} is an unseen pattern. 
%Secondly, there is some ``logical contamination'' in the dataset. That is, 2\% of sentences occur in both the training set and the validation/test set, modulo different variable names (e.g. $a \wedge \neg b$ and $d \wedge \neg e$). 
We leave this portion of the dataset unchanged, but in section~\ref{sec:generalization_methods} we will return to the topic of unseen patterns, and in fact use deliberate omissions to systematically test generalization performance.
%Thirdly, 

\paragraph{No unsatisfiability} A bigger shortcoming of the dataset is that there are no unsatisfiable sentences by design. %This makes the model harder to apply in practice.
The authors state: {\it``Entailment is a subproblem of satisfiability and could be encoded in the same form as our propositional formulas. (...) In contrast [to previous methods such as NeuroSAT] we apply a generic sequence-to-sequence model to predict the solutions to formulas, {\bf not only} whether there is a solution''} \citep{hahn_teachingtemporallogics_2021}.
However, the model cannot output whether there is a solution, it can only output a solution.
Arguably, a {\it wrong} prediction could be interpreted as the model's implicit way of predicting UNSAT, which is feasible because solution correctness can be checked quickly. We are aware that this limitation also limits the practical use of the models we train, but we have no easy fix and therefore proceed with a dataset without unsatisfiable formulas.\footnote{In preliminary experiments, adding unsatisfiable formulas to the training set (along with a special \mt{UNSAT} prediction token) did not significantly affect performance of the models trained on all data, nor the generalization capacities of models trained on the systematically different training sets.}
%We briefly reflect on the shortcomings listed above in \autoref{sec:futurework_data}.

\paragraph{Varying levels of difficulty} Lastly, since the sentences are generated randomly, they are not guaranteed to be difficult.
With a limited number of variables, there are also a limited number of correctly formatted possible outputs the model can generate.\footnote{This is in contrast to the trace generation task for Temporal Logic \citep{hahn_teachingtemporallogics_2021}. Traces can have arbitrary length, making a sequence-to-sequence setup more well-suited.}
A maximum of five variables means there are at most 242 possible outputs to choose from (of which $2^5=32$ are complete assignments and the rest are partial assignments). 
This makes the task considerably easier.\footnote{
In other propositional logic datasets, the number of variables can range from twenty to over a hundred \citep{evans_canneuralnetworks_2018,selsam_learningsatsolver_2019}.} 
%Randomly generated formulas are not guaranteed to be difficult to solve. 
For instance, formulas starting with a disjunction account for around 25\% of the dataset. 
These formulas are easier to guess a correct assignment for, as only one of its children has to be true for the entire sentence to be true.
For each formula in the validation set, we estimate its difficulty by calculating the relative number of satisfying complete assignments, or its ``possible worlds''.
As an example, for the sentence $a \wedge b$, only 1 out of 4 possible worlds is valid (\mt{a 1 b 1}), resulting in a valid assignment rate of 25\%.
The rate of valid assignments is high: 50\% on average for the validation set. A distribution can be seen in the Appendix in \autoref{fig:possible_worlds}(a). The majority of sentences (66\%) have a valid assignment rate of 40\% or higher. The number of {\it partial} assignments that are valid solutions is lower: 29\% on average.
Notably, longer formulas are not necessarily harder (\autoref{fig:possible_worlds}(b)). 
Given the lack of a straightforward solution, we proceed with the dataset as is, but emphasize that the accuracy scores we obtain must be assessed keeping this in mind. 

% Emacs, this is -*-latex-*-
%%% Local Variables:
%%% mode: latex
%%% TeX-master: "main"
%%% End:

\section{Reproduction and extension to other architectures}\label{sec:experimental_setup}

\subsection{Experimental setup}
\begin{figure}
  \centering
  \includegraphics[width=.3\textwidth]{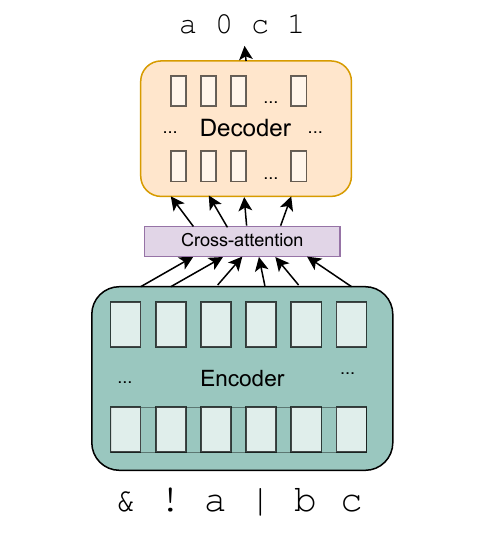}
  \caption{
  Illustration of the model architecture and task.
  The decoder is a standard Transformer model, for the encoder we experiment with three model architectures: Transformers, Graph Convolutional Networks, and LSTMs.
  }
  \label{fig:model_visualization}
\end{figure}

We use an encoder-decoder architecture as shown in \autoref{fig:model_visualization}. The encoder and decoder are both standard Transformer models \citep{vaswani_attentionallyou_2017}. The model's hyperparameters are based on the best model reported by \citet{hahn_teachingtemporallogics_2021}. 
Both the encoder and decoder have six layers, four attention heads, and a feedforward size of 512. We use an embedding size of 128 in the encoder and one of 64 in the decoder. In addition to self-attention within each layer, there are cross-attention weights from every layer in the decoder to the encoder's final hidden states in layer six, but not to earlier encoder layers. The  total number of model parameters is approximately 1.8 million. 
We vary the input positional encodings to be either absolute or based on tree structure \citep{hahn_teachingtemporallogics_2021,shiv_novelpositionalencodings_2019}.

\subsection{Other encoder architectures}
To compare Transformer models to other architectures, we replace the {\it encoder} of the model with a different architecture.

\paragraph{GCN}

We replace the Transformer encoder with a Graph Neural Network, specifically a Graph Convolutional Network (GCN) \citep{kipf_semisupervisedclassificationgraph_2017}. Each symbol in the input sentence is a node, that is connected to its children, its parents, and itself. %\todo{cite a paper that does this} \jj{I ChatGPT'd some related work and am finding some Chinese papers that do stuff vaguely in this direction, e.g. https://www.mdpi.com/2079-9292/14/3/423 But this is not cited often and might not be super relevant. Also this one: https://www.esann.org/sites/default/files/proceedings/2022/ES2022-88.pdf}. \anna{I'm sure I based this decision on a paper but I can't find it, i continue the search.} T
he architecture of the GCN explicitly encodes tree structure into the way the encoder processes its input. 
To keep the architecture similar to the Transformers, we simply replace the self-attention module by a GCN block. Thus, each encoder layer consists of a GCN block, followed by LayerNorm, followed by a ReLU activation. Residual connections are also kept between the layers. 
After a small hyperparameter search, we find that 16 encoder layers gives the best performance on the validation set. 
However, even with 16 layers, this model has around 900k parameters, which is less than both the Transformer and LSTM encoder variants.

\paragraph{LSTM}
We replace the Transformer encoder with a recurrent 6-layer LSTM-based encoder without attention. The decoder cross-attends to the final layer of the encoder. This allows us to test the effect of recurrence instead of self-attention. 
The initial hidden and cell state are learned. This model has 1.4M parameters in total, which is less than the Transformer models.
Neither the LSTM nor the GCN model make use of positional encodings.

\subsection{Training}%\todo{Move most of this to appendix?}
%All models are trained for 300k training steps (around 250 epochs) using a linear warmup scheduler with 4k warmup steps.
The training set contains 800K examples. The models are trained for 128 epochs, which is equivalent to 200k training steps with a batch size of 512. We use the Adam optimizer and 4000 warmup steps using the Noam learning rate scheduler. 
%\todo{lstm/gnn  different LR in a small hyperparam search.}
%
The model is trained using teacher forcing. At test time, we use greedy decoding.\footnote{Beam search is used in \citet{hahn_teachingtemporallogics_2021}, but, during preliminary experiments, a beam search with $K\in\{5,10\}$ yielded only very small improvements for our models (approximately $+1\%$ accuracy).}
The output of the model is all tokens before the first end-of-sequence token. 
We train at least two different model seeds for each combination of training set and model type, and report their average performance.

\subsection{Evaluation}
We report the performance of the base models on unseen data, both in terms of {\bf syntactic accuracy} (the model output exactly matches the ground truth output) and {\bf semantic accuracy} (the model output was a correct assignment, but not necessarily equal to the ground truth output). 

\begin{table}[h]
  \centering\sf
  \begin{tabular}[h]{l|l|l|l}
    \toprule
    Input & Ground truth output & Model output & Correctness   \\\midrule
    \texttt{\& !\ a | b c}& \texttt{a 0 b 1} & \texttt{a 1 b 0} & Incorrect \\
    \texttt{\& !\ a | b c}& \texttt{a 0 b 1} & \texttt{a 0 b 0 c 1} & Semantically correct \\
    \texttt{\& !\ a | b c}& \texttt{a 0 b 1} & \texttt{a 0 c 1} & Semantically correct \\
    \texttt{\& !\ a | b c}& \texttt{a 0 b 1} & \texttt{a 0 b 1} & Syntactically correct \\ \bottomrule
  \end{tabular}
  \caption{Evaluation of model outputs for a single example formula ($\neg a \wedge ( b \vee c )$). The model output can be semantically correct without exactly matching the given SAT solver output.}\label{tab:output_examples}
\end{table}

% Emacs, this is -*-latex-*-
%%% Local Variables:
%%% mode: latex
%%% TeX-master: "main"
%%% End:

%\section{Replication: Results}\label{sec:replication_results}
\subsection{Results}\label{sec:replication_results}
The main results can be seen in \autoref{tab:baseresults}. Our Transformer models achieve similar accuracy to \citet{hahn_teachingtemporallogics_2021}. Interestingly, syntactic and semantic accuracy on the {\it training} set is 57.99\% and 93.77\% respectively. %\jj{Would it be worth reporting train performance as well in the table then?}
These scores are similar to those of unseen data: this indicates that the models do not memorize the target assignments of the training data.\footnote{This holds for models trained on \mt{Prop35} as well as \mt{Prop35Balanced}.}
%\jj{What is the difference between validation and test data?}

Models are then tested on formulas of unseen lengths using the \mt{Prop50}-dataset \citep{hahn_teachingtemporallogics_2021}. We are able to partially reproduce the finding that tree positional encodings indeed help with length generalization. However, the semantic accuracy is lower overall.
This difference could at least partially be explained by the original dataset imbalance, which is amplified in \mt{Prop50}: the models of \citet{hahn_teachingtemporallogics_2021} may have a bias for parsing and predicting imbalanced trees.\footnote{Also see \autoref{subsec:imbalance_dataset}.}

The alternative architectures (GCN and LSTM) have slightly lower scores than the Transformer models. However, on out-of-distribution data, they perform better than the Transformer with absolute positional encodings. 

%\todo{Report on train set instead of val in table. Explain Hahn dagger.} Not feasible because don't have results for all models, takes long to evaluate. (will take hours if I need to evaluate every model, because train set is big and semantic accuracy has to be evaluated every time. )

%\todo{Possibly: LLM output?}

\begin{table}[!h]
  \centering\sf
  \begin{tabular}{l|rr|rr}
    \toprule
     &  \multicolumn{2}{c|}{\bf Prop35Balanced} & \multicolumn{2}{c}{\bf Prop50 (o.o.d.)}\\
    \bf Model     & Syntactic & Semantic & Syntactic & Semantic\\\midrule
    Absolute (\citeauthor{hahn_teachingtemporallogics_2021})  & - & - & (29.00) & (75.70)\\
    Tree (\citeauthor{hahn_teachingtemporallogics_2021})  &(58.10)$^\dagger$ & (96.50)$^\dagger$ & (35.80) & (86.10) \\
    \midrule
    Absolute (Ours) &\textbf{56.33}&\textbf{94.05}&24.07&66.17\\ 
    Tree (Ours)&54.03&91.03& \textbf{30.00} & \textbf{77.37} \\
\arrayrulecolor[rgb]{0.753,0.753,0.753}\midrule\arrayrulecolor[rgb]{0.0, 0.0, 0.0}
    GCN baseline& 50.52 & 86.98 & 26.74 & 72.30 \\
    LSTM baseline  & 50.30 & 89.66 & 25.22 & 74.72\\
    %Bottleneck, Absolute &55.71&93.24& 55.72&93.21&25.80&69.13\\ 
    %Bottleneck, Tree &53.16&89.65&53.06&89.63&29.25 & 75.22\\
    \bottomrule
  \end{tabular}

  \caption{Performance of all base models on unseen data. We report both syntactic and semantic accuracy. Semantically correct means that the model output is a valid partial assignment, whereas the output is syntactically correct only if the model output exactly matches the output produced by the symbolic solver. ($\dagger$): We report our reproduced accuracies on the test set of\mt{Prop35Balanced}; for comparison, we also give the scores from \citet{hahn_teachingtemporallogics_2021} on the original, unbalanced test set \mt{Prop35} as listed in their paper (reported between brackets). Note that results on the unbalanced data may overestimate the learning abilities of the networks.
  }\label{tab:baseresults}
\end{table}

%All models, including the GCN and LSTM, reach at least 97\% semantic accuracy on the templated test set.
%\jj{Dit moet je me nog maar even uitleggen hoe dat zich verhoudt tot Table 3}

%%% Local Variables:
%%% mode: LaTeX
%%% TeX-master: "main"
%%% End:

\section{Generalization to Unseen Patterns: Methods}\label{sec:generalization_methods}
Up to this point, we have successfully replicated the Hahn et al.\ results with Transfomrers, and  extended their work to other deep learning architectures, while training on the entire train set and reporting aggregate accuracies over the entire test set. In this section, we consider the problem of how to give a more finegrained assessment of the type of generalizations that the trained models have learned or can learn.

The previous section already considered one aspect of generalization: length generalization (sometimes called {\it productivity}). It tests how well models understand inputs (and in the context of sequence generation, outputs) that are longer than those seen during training \citep[e.g.][]{evans_canneuralnetworks_2018,hupkes_compositionalitydecomposedhow_2020,selsam_learningsatsolver_2019}.
As seen in \autoref{sec:replication_results}, our models (perhaps with the exception of the Transformer with absolute positional encodings) perform reasonably well when tested on \mt{Prop50}: they show productive behaviour.
The formulas in \mt{Prop50} provide some syntactic challenge - they are longer and may be harder to parse. However, the data is generated according to the exact same distribution, meaning statistical artifacts remain \citep{zhang_paradoxlearningreason_2022}.
There is also no distribution shift in outputs -- as discussed in \autoref{sec:dataset_issues}, valid assignment rates in \mt{Prop50} are similar to the original \verb\Prop35\, despite the change in length.
%For other
In those terms, the dataset is not necessarily ``harder'', either.\footnote{Note that for some logic datasets, such as ones that are constrained to Conjunctive Normal Form (CNF), there {\it does} exist a strong correlation between length and possible-world difficulty -- which is why it is more impressive for models trained on this data to generalize productively \citep{selsam_learningsatsolver_2019}.}

Therefore, a more interesting way to measure the generalization capacities of models is to test the model on combinations of tokens that are unseen in the training phase. This type of generalization is called {\it systematic} generalization \citep{lake_generalizationsystematicitycompositional_2018,hupkes_compositionalitydecomposedhow_2020,loula_rearrangingfamiliartesting_2019}. 
If the model has truly learned the semantics of propositional logic, it should be able to combine these unseen structures, and its performance should not, or barely, decrease.

Since all possible combinations of tokens\footnote{Except for double negation (\mt{!\ !}).} are present in \mt{Prop35}, we design seven alternative training sets.
For each new training set, exactly one pattern is left out of the model training data. Crucially, we do use the pattern at test time, to assess the extent to which the trained models can generalize to this unseen pattern. 
The set of patterns for the generalization splits is listed in \autoref{tab:unseen_patterns}.
The variety of patterns allows us to test for multiple kinds of systematic generalization. 
\begin{table}[!h]
  \centering\sf
  \begin{tabular}{rllr}
    \toprule
    \# & Pattern & Description & \% \\\midrule
    (P1)& \mt{!\ \&}& A negated {\sc and}-node & 41.06\\
    (P2)& \mt{!\ |}& A negated {\sc or}-node&40.96\\
    (P3)& \mt{!\ xor}& A negated {\sc xor}-node&24.56\\
    (P4)& \mt{!\ b}& A negated $b$-node. & 47.42\\\midrule
%    (P5)& \mt{\& !\ A ! B}& An {\sc and}-node with two negated subtrees&18.60\\
    (P5)& \mt{\& !}& An {\sc and}-node with a {\it left} child that is negated & 52.07\\\midrule
    (P6)& {\color{Highlight}\mt{\& xor}} \mt{\_ C}, {\color{Highlight}\&}\mt{ C }{\color{Highlight}xor} \mt{\_}& An {\sc and}-node with a child that is a {\sc xor}-node & 25.93\\
    (P7)& {\color{Highlight}\mt{<-> !\ }}\mt{A B}, {\color{Highlight}\mt{<->} }\mt{A }{\color{Highlight}!\ } \mt{B}& An {\sc iff}-node with a child that is negated & 49.21\\
    \bottomrule
  \end{tabular}
  \caption{The unseen patterns, their description, and their overall occurence in the original training data. All unseen patterns consider some form of direct parent-child combinations. The first four patterns test the compositionality of the {\sc not}-operator (\mt{!}). Patterns (P5) tests unseen types of trees, and patterns (P6) and (P7) test the compositionality of operators \mt{\&} and \mt{<->}. Some combinations of operators may be more challenging than others.}
  \label{tab:unseen_patterns}
\end{table}

The first four patterns all assess the genericness of the {\sc not}-operator. However, they may not be equally difficult. For instance, a model trained without pattern (4) may still be able to perform well on it, as it still needs to learn to apply the {\sc not}-operator to four other distinct variables during training. Another aspect is that for any model, a formula starting with the pattern \mt{!\ \&} is easier to guess correctly compared to \mt{!\ |},  since on average, there are more worlds in which the former formula is true.
All first four patterns include completely unseen sequences of symbols in the flat string. For instance, a \mt{!}-token is never followed by a \mt{|}-token for pattern (1). 
Pattern (5) tests an unseen flat string in the input: \mt{\&} is never followed by \mt{!}. The {\sc and}-operator can however still be seen in combination with a NOT-subtree, but only if that subtree is its righthand child. 
Finally, patterns (6) and (7) test for the compositionality of the {\sc and}-operator and the {\sc iff}-operator, in the same way that the first four patterns did for the {\sc not}-operator. Leaving out {\sc and} may be less of an impairment than leaving out {\sc not} or {\sc iff}: {\sc and} is fully compositional in the sense that the truth values of its subtrees can stay intact upon combination.
%\footnote{Four additional generalization splits were tried: these all had similar results to patterns (5) through (8), and were not included in the main results. Details and results can be found in \autoref{sec:appendix:additional_generalization_splits}.}

It is important to note that omitting any of these patterns does not reduce the semantic expressivity seen during training, as formulas containing these patterns can be expressed as differently structured sentences with the exact same possible worlds.
Similar to the productive generalization split, the held-out data is not ``harder'' in terms of possible correct outputs. 
This allows us to test for systematic generalization without also changing the output distribution: the outputs do not become longer, more complex, or harder to guess: it is only the input that changes, thus, we are able to test for true structural generalization.
%Finally, the well-defined semantics of propositional logic also allows us to keep the {\it amount} of training data intact without having to generate new data.

%\todo{One other aspect that I'm not describing here is there is change in the input in terms of token appearance. Some datasets do control for this.}

%We take care 
%\todo{Synthesize a sentence about that the number of worlds does not decrease for these patterns either, but that they are still more interesting than simple length splits.}

\subsection{Details on the creation of different splits}
We create these splits by either rewriting the formulas in the existing dataset to a logically equivalent formula, or by simply removing sentences that include the pattern. The rewriting approach has the advantage of maintaining the same training set size, including the ground truth outputs, without the need to generate more data. Any double negations (\texttt{!\ !}...) that occur during this rewriting process are removed, as they are not present in the original dataset.

%\todo{We call this 'semantic consistency', o.i.d. ?
%Problem: I currently don't do this with the final two patterns, and also report results with pattern P4 without semantic consistency, shifting the output distribution. However: I balanced the dataset in advance, which also might change outputs (1 in 2)  -- hence, rewriting does not necessarily hinder model understanding.}

Specifically, using the DeMorgan's laws, pattern (P2) is rewritten into pattern \mt{\& !\ A !\ B}, and
pattern (P1) is be rewritten to \mt{| !\ A !\ B}. 
Pattern (P3) is equivalent to \mt{<->}, and can also be rewritten.
For pattern (P4), we rewrite any occurrences of \mt{\& !\ A !\ B} with \mt{!\ | A B}, and occurrences of \mt{\& !\ A C} with \mt{\& C !\ A}. 
%For pattern (6), we start with the rewritten dataset for pattern (5), as it is a superset of this pattern. Now that we know that no \mt{\&}-node has two negated children, we replace any remaining occurences of \mt{\& !\ A B} with \mt{\& B !\ A}.
%
%For pattern (4), we rewrite the pattern to \mt{!\ \& b b}, to avoid a shift in the output distribution. \footnote{This pattern is not present in the original dataset, which could affect the model's behaviour. However, when removing all sentences containing \mt{!\ b}, only 30\% of outputs containing \mt{b 0} remains. This could bias the model to predict \mt{b 1} too often, as the original truth value outputs are balanced per variable. We therefore do include the pattern \mt{!\ \& b b} for this case. In preliminary experiments, we found that a model trained on data where any sentence containing \mt{!\ b} were removed, still had an acceptable performance of 84\% semantic accuracy on sentences containing \mt{!\ b}.}
%
For pattern (P4) and the final two patterns (P6) and (P7), we simply remove them from the training data.

\subsection{Templated test set}
%We test our models on the existing validation and test sets, as well as a templated dataset containing short sentences.
As described in \autoref{sec:dataset_issues}, longer formulas are not necessarily harder to solve.
Furthermore,  it is difficult to gauge how relevant a subpattern in a randomly generated formula is. For instance, to solve a formula with \texttt{or} as the main connective, the model may ignore half of the formula, which makes it harder to interpret its behaviour. 
To address this, we also test all models on a test set of shorter sentences constructed based on templates.

Using templated data, we have some control over whether a model needs to understand(/solve) the subpattern in order to understand(/solve) the whole sentence. 
%\todo{This is relevant for semantic generalization. Long sentences are more of a syntactic challenge, but longer sentences do not necessarily.}
%
For a set of 12 simple patterns (based on the unseen patterns), we generate sentences with the pattern in a number of contexts, to systematically test the model ability to correctly interpret the pattern in these contexts. 
For instance, to test whether models understand the pattern $\neg (A \oplus B)$ (negation applied to the xor-operator) in different contexts, we generate for instance at least the following sentences:
\begin{itemize}
\item {\color{Highlight}$\neg$}$(a ${ \color{Highlight}$\oplus$ }$ c)$ (Basic understanding)
\item {\color{Highlight}$\neg$}$(\neg e${ \color{Highlight}$\oplus$ }$\neg c)$ (Applied to negated variables)
%\item $\neg ((a \wedge e) \oplus (b \oplus c))$ (Applied to operators)
\item $(${\color{Highlight}$\neg$}$(a${ \color{Highlight}$\oplus$ } $c)) \wedge e$ (Understanding in context of an \texttt{and})
\item $(${\color{Highlight}$\neg$}$(a ${ \color{Highlight}$\oplus$ }$ c)) \leftrightarrow (\neg e)$ (Understanding in context of an \texttt{iff})
\item $(${\color{Highlight}$\neg$}$(a ${ \color{Highlight}$\oplus$ }$ c)) \vee (\neg e \wedge e)$ (Understanding in context of an \texttt{or}, the other half of the subtree is unsatisfiable)
\item \dots~(Combinations of the above, see Appendix~\ref{subsec:templated_dataset})
\end{itemize}

A full description of the templated test set is listed in Appendix \ref{subsec:templated_dataset}. 
Note that the templated dataset may contain instances seen during training.
All models, including the GCN and LSTM, get at least 98\% accuracy on the templated dataset.

%%% Local Variables:
%%% mode: latex
%%% TeX-master: "main"
%%% End:

%\jj{This section header could be a bit more descriptive}

\section{Generalization to Unseen Patterns: Results}\label{sec:generalization_results}

%\subsection{Systematic Generalization}

\begin{figure}[h]
  \centering
  \includegraphics[width=.9\textwidth]{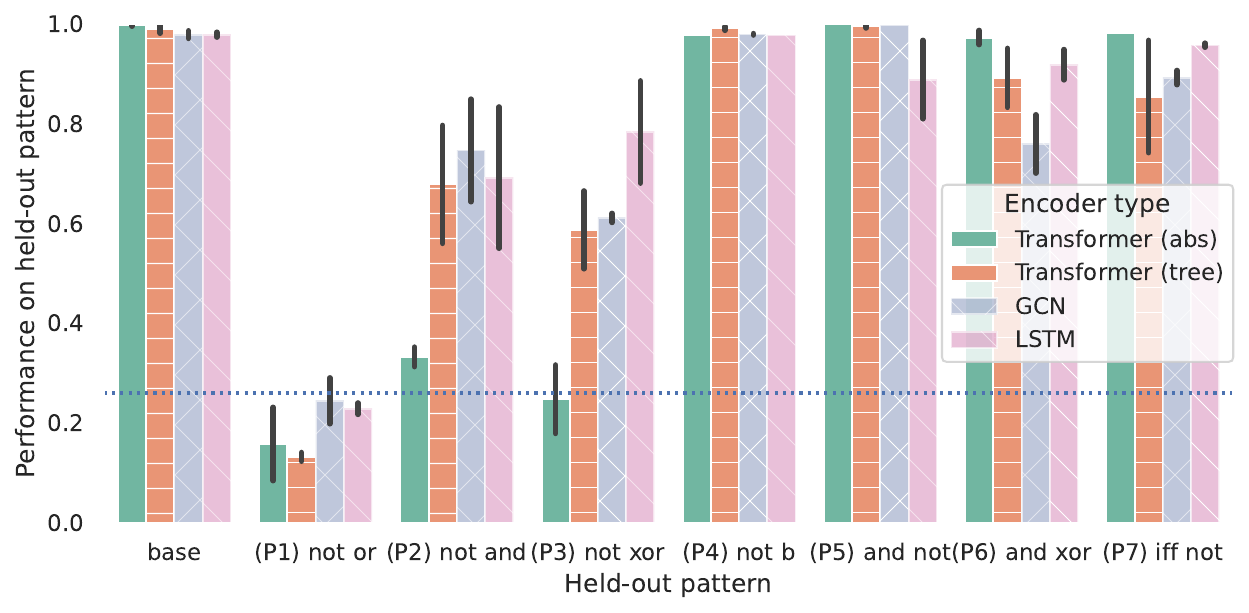}
  \caption{Generalization on the templated test set, for all architectures. Each model (except the base models) is evaluated on sentences containing its respective held-out pattern. Standard error shown is between model seeds. The dotted line shows the partial assignment random guessing chance on the templated test set (26\%).}\label{fig:generalization}
\end{figure}

We now turn to the evaluation of the models trained on systematically different data. All seven models are evaluated on the same (original) validation set. We report performance on relevant subsets of the validation set. We will refer to models trained on the original training set, that was evaluated in \autoref{sec:replication_results}, as the {\it base} models.
All base models, including the GCN and LSTM, reach at least 97.5\% semantic accuracy on the templated test set. The results for all other models are summarized visually in \autoref{fig:generalization}: each model is evaluated on a subset of the validation set that contains the unseen pattern. Importantly, the performance for {\it all} models stays equal on sentences where the unseen pattern is {\it not} present.

\paragraph{Transformer models}
When comparing Transformer models trained without patterns (4) through (7) to the base model, the difference in performance on the unseen pattern is relatively small, or in some cases nonexistent. This indicates that these unseen patterns can be structurally combined at test time.

In contrast, there is a significant drop in performance when either of patterns (1), (2) or (3) are left out of training. These are all patterns that test the generalizability of the {\sc not}-operator. Whereas models trained without pattern (4) are able to combine negation with an unseen {\it variable}, these models are not able to combine negation with an unseen {\it operator}.
Sentences that do not contain the unseen pattern are unaffected, yielding scores similar to the base models.

Tree positional encodings help systematic generalization for (P2) and (P3), but performance on these patterns still lags significantly behind compared to the rest of the patterns.
%\todo{Group this figure by model?}

\begin{table}
  \centering
  \begin{tabular}{ll|l|lc}
    \toprule
    \bf Pattern & \bf Model type(s) & \bf Input & \bf Output & \bf Correct?\\\midrule
    (P1) not or & Transformer-tree & $\neg (a \vee b)$ & \texttt{a 1 b 0} & N \\
    (P1) not or & GCN & $\neg (e \vee c)$ & \texttt{c 1} & N \\
    (P1) not or & LSTM & $\neg (e \vee c)$ & \texttt{e 0 } & N \\\midrule
    (P2) not and & Transformer-abs & $\neg (e \wedge b)$ & \texttt{b 1 e 1} & N\\
    (P2) not and & Transformer-tree, LSTM & $\neg (e \wedge b)$ & \texttt{b 0 e 1} & Y\\
    (P2) not and & GCN & $\neg (e \wedge b)$ & \texttt{b 0 e 0} & Y\\
    \midrule
    (P3) not xor & Transformer-tree, Transformer-abs & $\neg (a \oplus b)$ & \texttt{a 1 b 0} & N \\
    (P3) not xor & LSTM, GCN & $\neg (a \oplus b)$ & \texttt{a 1 b 1} & Y \\
    (P3) not xor & Transformer-tree, GCN & $\neg ((a \oplus b) \oplus e)$ & \texttt{a 1 b 1 e 1} & N \\
    (P3) not xor & LSTM & $\neg ((a \oplus b) \oplus e)$ & \texttt{a 1 b 1 e 0} & Y \\
    \bottomrule
    %    (P3) not xor & trafo-tree & $\neg (a \oplus c)$ & \texttt{b 0 e 0} & Y\\
    %(P3) not xor & gcn & $\neg ((a \vee b) \oplus e)$ & \texttt{a 1 e 1} & Y\\\midrule
    %(P3) not xor & gcn & $e \oplus (\neg (a \oplus b))$ & \texttt{a 1 b 1 e 1} & N\\
    %(P3) not xor & lstm & $\neg (a \oplus c)$ & \texttt{b 0 e 0} & Y\\

  \end{tabular}
  \caption{Outputs for some simple formulas including the first 3 held-out patterns.}\label{tab:qualitative_samples}
\end{table}

\paragraph{Other architectures}
The GCN and LSTM models also perform relatively well on patterns (P4) through (P7). Surprisingly, the GCN architecture lags behind on pattern (P6). Its accuracy on sentences containing the pattern \textsc{and xor} is 76\%, its accuracy on sentences starting with this pattern is even lower, 57\%.

None of the architectures reliably solve pattern (P1). None of the models produce correct outputs for the simplest sentences for pattern (P1), although the models seem to produce different wrong answers, as shown in \autoref{tab:qualitative_samples}.

%\todo{Mention inductive bias again.}

Both the GCN and LSTM show some improvement over Transformers on pattern (P2), and the LSTM outperforms all models on pattern (P3).

% \subsection{Detailed results}
% In 

\begin{figure}[!ht]
  \centering
  \includegraphics[width=\textwidth]{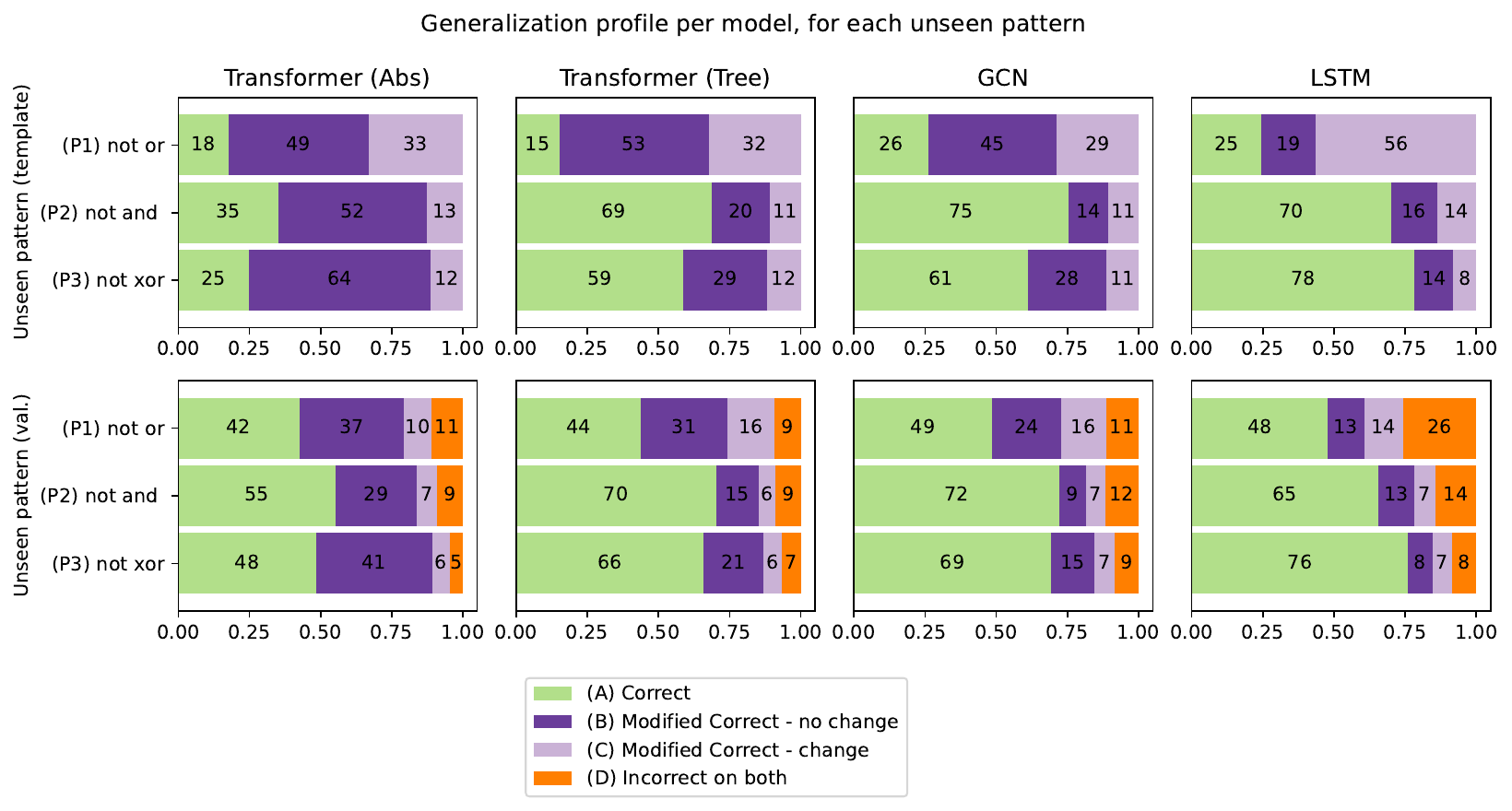}
  \caption{Unchanged predictions, per model, per pattern. Results are shown on the templated test set (top row) and PropRandom35 validation set (bottom row). `Modified' denotes the sentence where the relevant pattern is non-negated. When the model prediction is the same there as for the original sentence, we can say that the model is ignoring the negation operator.}\label{fig:unchanged}
\end{figure}

\subsection{Behavioural results}

\paragraph{Are non-generalizing models simply ignoring the {\sc not}-operator in unseen combinations?}
We test whether a model ignores the negation-operator by replacing any occurences of $\neg \phi$ with $\phi$, and checking if the model behaviour changes on this new input.

At this point, we classify model behaviour into four categories. After introducing the negation operator, the model either:
\begin{enumerate}[(A)]
\item Correct(ly changes its prediction)\footnote{Note that correct possible worlds for the original and modified sentence may overlap. In this case it is not necessary to change the prediction, the result will be counted as correct.} -- correctly processing the negation-operator
\item Does not change its prediction (ignores the negation-operator)
\item Changes its prediction to an alternative output for the original sentence (may be ignoring the negation-operator)
\item Changes its prediction to something else
\end{enumerate}

\autoref{fig:unchanged} shows an overview of these four classes of model behaviour for patterns (P1), (P2) and (P3).
Transformers with absolute encodings show the most `ignoring' behaviour for all unseen patterns. The GCN and LSTM baselines are the least likely to ignore the negation operator.

\paragraph{Are other outputs of non-generalizing models affected?}
All models perform well on subsets of the validation set that do not include the held-out pattern. We now investigate whether these (correct) predictions are {\it different} to those of other models.

All base models have around 83\% overlap in their predictions, meaning only 17\% of predictions are different between different seeds of the same architecture.

Predictions produced by the seven models with alternative training sets do not differ more than 18\% from the base model or from one another, when only looking at those predictions on sentences that do not include the relevant pattern.
When only comparing predictions for the subset of sentences that contain the relevant binary operator, there are no changes either.
Thus, in terms of model behaviour, individual operators are not affected by the modified training set.

%\subsubsection{Generalization during early training}
% Emacs, this is -*-latex-*-
%%% Local Variables:
%%% mode: latex
%%% TeX-master: "main"
%%% End:

\section{Discussion and Conclusion}\label{sec:discussion}
We briefly reflect on our findings, and present an outlook for future work.
We examined whether neural models trained on limited data learn the semantics of propositional logic. 
We designed several generalization splits based on semantic consistency. 

When trained on i.i.d. data, Transformer models are able to produce novel correct outputs to unseen sentences with an accuracy of 94\%. 
When replacing the Transformer encoder with a GCN or LSTM, these models still obtain an accuracy of 87\%.\footnote{We cannot completely exclude the possibility that the lower accuracy of GCNs and LSTMs is due to a relatively limited hyperparameter optimization for these models. We did our best to find the best possible settings, but only for the Transformer models we could make fruitful use of the hyperparameters found by \citet{hahn_teachingtemporallogics_2021}.}

Even when trained on structurally divergent data, these models demonstrate remarkable flexibility, even without specialized tree positional encodings.
They are able to combine novel combinations of operators (such as (P7), in which the models are able to apply bi-implication to negated nodes ($\leftrightarrow$ to $\neg$), while never having seen the combination in this order) and variables (such as (P4), in which the models never saw $\neg$ applied to the variable $b$).
The ability of the Transformers to represent most operators in a way that they can be combined at test time is one indication of the learnability of systematic compositionality. Another indication is their capacity to generically represent variables as {\it instances} of the same type \citep{marcus_algebraicmindintegrating_2003}.

Interestingly, however, models do struggle on generalizing to unseen \textit{negated operators}.
% The only exception is when the operator that needs to be systematically combined is {\it negation}. 
Models that were trained without \mt{!\ \&} (P1), \mt{!\ |} (P2), or \mt{!\ xor} (P3) all failed to apply negation to the binary operator when the pattern is reintroduced at test time.
%Using tree positional encodings for these models did not improve their systematic generalization on all these patterns, indicating that the issue is not just one of syntactic confusion (failure to parse).
Adding explicit structure to the encoder architecture can improve performance for some patterns (such as (P2) and (P3)), but not all. 
Notably, the LSTM-based models also show improvements over a standard Transformer on these unseen patterns, even though they do not have access to the underlying graph structure, suggesting that recurrence is useful as an inductive bias for systematic generalization.
None of the architectures trained without the pattern ``not or" can solve the simple sentence $\neg (a \vee b)$.

%\todo{LSTMs perform well even though they do not have explicit information about the input structure AND even though their hyperparameters were not tuned extensively. This .... INDUCTIVE bias.}

Even though the generalization performance improves for these architectures, their performance on patterns with negation does not match the generalization performance on patterns that do not include a negated operator. 
This suggests that while Transformer models are able to learn generalizable representations for variables and binary operators, they fail to form a unified representation of negation.
This issue extends beyond our logic-based tasks: modern LLMs, such as those from the Llama-3 or GPT-4 family, notoriously struggle to understand linguistic negation as well \citep{kassner_negatedmisprimedprobes_2020,truong_languagemodelsare_2023,she_sconebenchmarkingnegation_2023,wan_triggeringlogical_2024}.
The failure to learn a generalizable negation operator may suggest that not all types of systematicity can be learned by Transformers simply using next-word prediction and backpropagation.
A solution to this problem may be found in alternative optimization strategies, incorporating reinforcement learning, diffusion, and neurosymbolic methods in the training of our models \citep{DBLP:conf/iclr/YeGGZJLK25, yolcu_learninglocalsearch_,vankrieken2025neurosymbolicdiffusionmodels}. %mccoy_embersautoregressionshow_2024}.
% \todo{Therefore, we might want to go beyond autoregression and towards neurosymbolic methods to solve these problems \citep{ye_autoregressiondiscretediffusion_2025,mccoy_embersautoregressionshow_2024}.}

% \todo{Recap from the intro about the latest LLM/reasoning problems: Lessons learned, but this does ofc not solve the greater problem.}
% \todo{The bigger picture: How do we test the generalisation of reasoning models? Interpretability?}

\paragraph{Outlook}
We started our introduction citing recent work on reasoning in Large Language Models and Large Reasoning Models (LRMs); are there any lessons from our detailed case study of propositional logic learning to how we evaluate these models?
One possible route is to use the variable assignment task we studied in this paper as just another "logic puzzle" to be added to the already long list of reasoning tasks that are included in benchmarks for these large models. This seems to be the approach in, for instance, LogicBench \cite{parmar-etal-2024-logicbench}. 
But such approaches often end up, again, with summarizing the performance of models with a single score, aggregated over all the puzzles. One of the lessons from our case study was that aggregate scores hide much of the interesting differences between models.

Another possible route would be to apply our methodology of systematically leaving out specific patterns from the training set to test for true generalization.
But such an approach runs into major computational challenges: training LRMs is so costly, that it is unfeasible to train 7 different models just to obtain insights into their generalization behavior.

How may we then systematically test generalization behavior of LRMs on logic puzzles, without retraining these enormous models? 
We suggest a possible third route is to study small and large model side by side, and use the small models as surrogate models to \emph{predict} the generalization behavior of large models. 
Such an approach would need a variety of candidate surrogate models, and a variety of techniques from the field of interpretability to investigate which surrogate model best approaches the learned solutions for a specific task in the large models.

For propositional logic variable assignment, the candidate surrogate models could, perhaps, include the 4 models we have presented in our case study; to measure how similar their model internals are to large models, techniques such as Centered Kernel Alignment \cite{kornblith19iclr_CKA} and automatic circuit discovery \cite{conmy23neurips} could prove useful.
We leave such exciting new research directions for future work.

%\todo{Using well-defined tasks and generalization splits based on semantic consistency... }

%%% Local Variables:
%%% mode: latex
%%% TeX-master: "main"
%%% End:

% \section{Conclusion}\label{sec:conclusion}
% \input{conclusion}

% Please only edit 'Additional'  when adding references, EverythingResearch gets overwritten by Zotero every time I download a new paper
\bibliographystyle{ACM-Reference-Format}
\bibliography{bib/EverythingResearch,bib/Additional}

\appendix
\section{The Dataset}

\subsection{Imbalance in the dataset}\label{subsec:imbalance_dataset}
Preliminary inspection of the dataset showed that the generated formulas in \mt{Prop35Random} are imbalanced in one direction: left subtrees had an average size of 2.7 tokens, whereas right subtrees were larger, containing 5.5 tokens on average.
This harms the ability of the model to solve left-imbalanced trees. In a preliminary experiment, we test models trained on right-imbalanced data on inverted trees (that are left-imbalanced -- inverting trees is possible because of the aforementioned commutativity of all binary operators used), which results for drops in accuracy for both absolute and tree positional encodings, as can be seen in \autoref{tab:performance_imbalanced}.

\begin{table}[h]
  \centering
  \begin{tabular}{l|rrrr}\toprule
    {\bf Model (training set)} & \multicolumn{4}{c}{\bf Performance on validation set}\\
     & Prop35 & Prop35-Inverted & Prop35Balanced & Prop50  \\\midrule
    Abs. (Prop35)& 94.5 & 56.6 & 71.2 & 73.7\\ 
    Tree (Prop35)& 93.7 & 68.5 & 81.2& 80.3\\\midrule
    Abs. (Prop35Balanced)& 94.5 & 94.6 & 94.6 & 67.9\\
    \bottomrule
  \end{tabular}
  
  \caption{Performance (semantic accuracy) of models trained on right-imbalanced data (first two rows) versus models trained on balanced data (third row). The right-imbalanced models have an advantage on right-imbalanced Prop50, but do not perform well on balanced data.}
  \label{tab:performance_imbalanced}
\end{table}

To mitigate the subtree imbalance, we rebuild the dataset by randomly flipping 50\% of the subtrees to create \mt{Prop35Balanced}. \autoref{tab:balanced} shows the average subtree sizes before and after balancing.
We keep the target sequences the same, as they are still semantically correct.
Note that the symbolic solver may give a different assignment out of all possible assignments.\footnote{This happens in about 50\% of datapoints in the validation set.}
The model now does not have access to the exact symbolic solver output for a formula, which may improve its semantic understanding, but hurt its ability to emulate the solver.
Our results from \autoref{tab:performance_imbalanced} however show that there is no drop in accuracy, suggesting that rewriting the data, even in a way that would change ground truth outputs, does not harm logical understanding. 

\begin{table}[h]
  \centering\sf
  \begin{tabular}{l|rr|rr|rr}
    \toprule
    & \multicolumn{2}{c|}{\texttt{Prop35}}& \multicolumn{2}{c|}{\texttt{Prop50}} & \multicolumn{2}{c}{\texttt{Prop35Balanced}} \\
    & Left & Right & Left & Right & Left & Right \\\midrule
    Root node sizes & 6.12& 14.03& 14.7 & 26.7 & 10.07 & 10.08\\\midrule
    All node sizes & 2.73& 5.46& 3.7 & 6.9 &    4.05 & 4.05\\
    \bottomrule
  \end{tabular}
  \caption{Average subtree sizes of the training set, before and after balancing.}
  \label{tab:balanced}
\end{table}

It is important to note that imbalanced trees still exist in \mt{Prop35Balanced}, just not in (overwhelmingly) one direction. Models trained on this new dataset perform equally well on balanced trees and imbalanced trees: i.e., both the original imbalanced validation set, and the inverted imbalanced validation set. %\todo{Is syntactic accuracy the same too?}

\subsection{Difficulty of the dataset}
\begin{figure}[!h]
  \centering
  \includegraphics[width=0.49\textwidth]{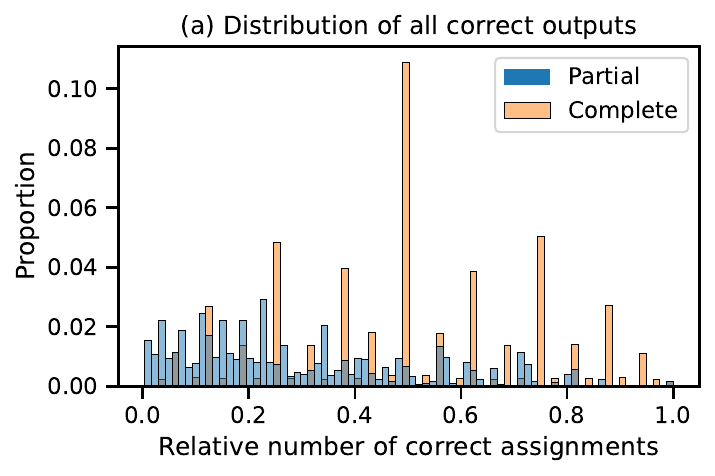}
  \includegraphics[width=0.49\textwidth]{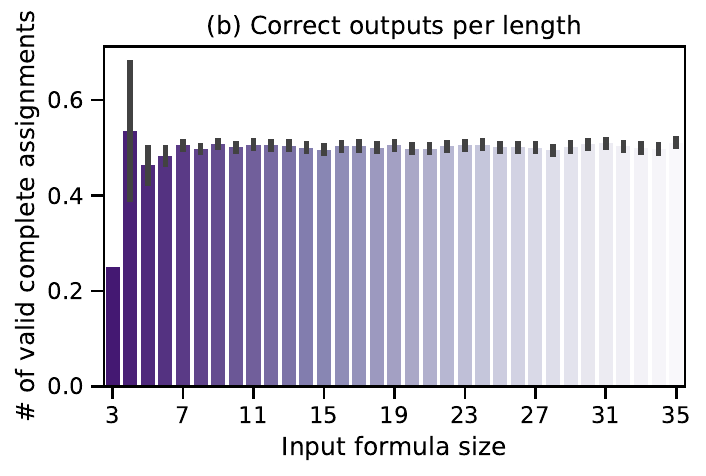}
  \caption{Distribution of possible worlds in the validation set. }
  \label{fig:possible_worlds}
\end{figure}

\autoref{fig:possible_worlds} shows that length of the input formula does not correlate with the difficulty of the output in terms of possible correct outputs.

\clearpage \newpage
\subsection{The templated dataset}\label{subsec:templated_dataset}
We use a set of 17 templates. We use \texttt{E} as a placeholder for subformulas.

\begin{enumerate}
\item \makebox[2cm][l]{\texttt{E }} The formula itself.
\item \makebox[2cm][l]{\texttt{\& E e        }} Ability (simple) to process in $\wedge$-root, on the left side
\item \makebox[2cm][l]{\texttt{\& ! e E      }} Ability (simple) to process in $\wedge$-root, on the right side
\item \makebox[2cm][l]{\texttt{<-> E e      }} Ability (simple) to process in $\leftrightarrow$
\item \makebox[2cm][l]{\texttt{| E \& e ! e  }} Ability (simple) to process in $\vee$-root (other half is UNSAT)
\item \makebox[2cm][l]{\texttt{xor e E      }} Ability to process in $\oplus$-root, on the right side
\item \makebox[2cm][l]{\texttt{<-> E e      }} Ability to process in $\leftrightarrow$-root, on the left side
\item \makebox[2cm][l]{\texttt{! xor E e    }} Nested $\neg \oplus$
\item \makebox[2cm][l]{\texttt{! xor ! e E  }} Nested $\neg \oplus$, 2
\item \makebox[2cm][l]{\texttt{! \& ! e E    }} Nested $\neg \wedge$
\item \makebox[2cm][l]{\texttt{! \& e E      }} Nested $\neg \wedge$, 2
\item \makebox[2cm][l]{\texttt{! | E e      }} Nested $\neg \vee$
\item \makebox[2cm][l]{\texttt{! | ! e E    }} Nested $\neg \vee$, 2
\item \makebox[2cm][l]{\texttt{\& ! E e      }} Nested $\wedge \neg$
\item \makebox[2cm][l]{\texttt{\& e xor E e  }} Nested $\wedge \oplus$
\item \makebox[2cm][l]{\texttt{\& xor E ! e e}} Nested $\wedge \oplus$, 2
\item \makebox[2cm][l]{\texttt{<-> ! E e   }} Nested $\leftrightarrow \neg$
               
\end{enumerate}

We then define 13 possible subformulas $\texttt{E}$. They include all the possible held-out patterns. The variables \texttt{A} and \texttt{B} are used as placeholders for actual variables or small formulas, where  \texttt{A} $\in\{$  \texttt{a}, \texttt{e}, \texttt{!e}, \texttt{\& a e}, \texttt{! \& a e}, \texttt{| !e !a} $\}$ and \texttt{B} $\in\{$\texttt{b}, \texttt{c}, \texttt{!c}, \texttt{\& c d}, \texttt{xor b c}, \texttt{<-> c b}$\}$.

\begin{enumerate}[(i)]
\item \texttt{! xor A B}
\item \texttt{! <-> A B}
\item \texttt{! | A B}
\item \texttt{! \& A B}
\item \texttt{xor A B}
\item \texttt{<-> A B}
\item \texttt{| A B}
\item \texttt{\& ! A B}
\item \texttt{\& ! A ! B}
\item \texttt{\& xor A B c}
\item \texttt{\& c xor A B}
\item \texttt{<-> ! B A}
\item \texttt{<-> A ! B}
\end{enumerate}

For example, template (14) with chosen \texttt{E} = subformula (vii) yields 36 possible formulas:
\begin{itemize}
\item \texttt{\& ! | a b e}
\item \texttt{\& ! | a c e}
\item \texttt{\& ! | a ! c e}
\item \texttt{\& ! | a \& c d e}
\item \dots
\item \texttt{\& ! | ! a <-> c b e}
\end{itemize}

We generate all possible combinations of \texttt{E}, \texttt{A} and \texttt{B} for each template. For each resulting formula containing $\neg \psi$ ($\psi \in \{\vee,\wedge,\oplus\}$), we also add the formula in which instances of $\neg \psi$ are replaced with $\psi \neg$.\footnote{\ This allowed us to run additional behavioural tests to see whether models trained without $\neg \psi$ systematically produce (incorrect) answers that correspond to the modified formula containing only $\psi \neg$, instead. After preliminary experiments, however, during preliminary experiments, we find that the `ignore' hypothesis (results in \autoref{fig:unchanged}) is a better explanation of model behaviour in terms of systematic mistakes in the output.} If a resulting formula contains a double negation or is unsatisfiable, we remove it. We remove any duplicate formulas. The final templated test set contains 8301 possible formulas.

\subsection{Detailed experimental setup}
For all Transformer and GCN models, we use a learning rate of $1e^{-4}$. For the LSTMs, we use a learning rate of $5e^{-5}$.

 For the Transformer models trained with absolute positional encodings and the LSTMs, every input sentence is wrapped between two extra tokens indicating the beginning-of-sequence and end-of-sequence.

The outputs for all models are followed by an end-of-sequence token.
\clearpage\newpage
\section{Detailed results}

\begin{figure}[h]
  \centering
  \includegraphics[width=.48\textwidth]{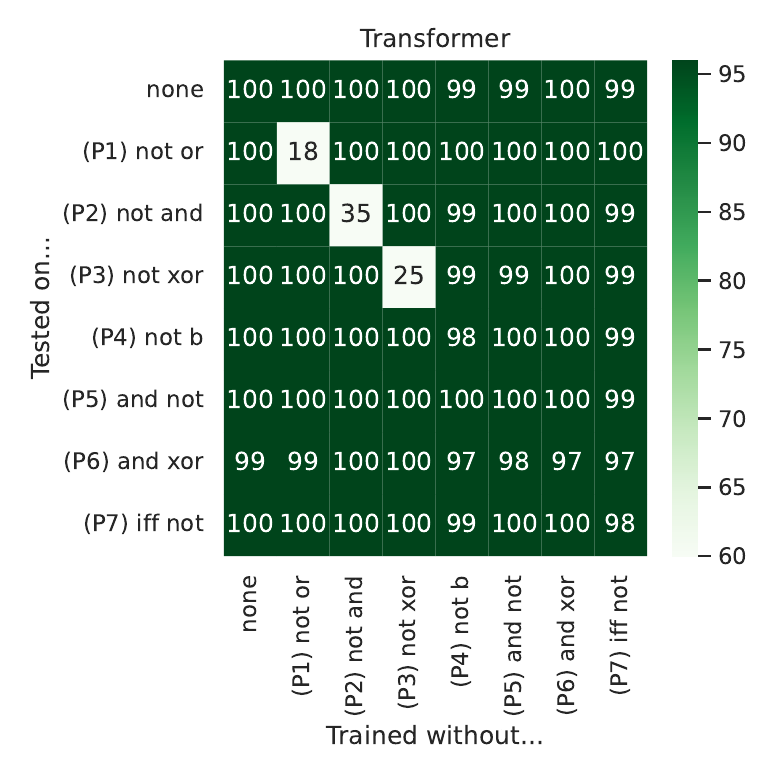}
  \includegraphics[width=.48\textwidth]{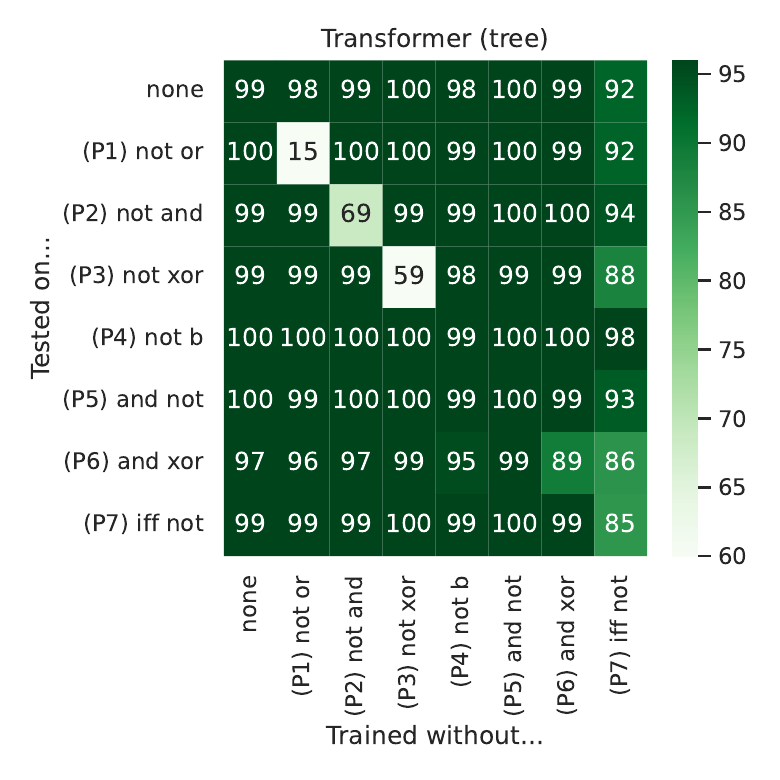}
  \includegraphics[width=.48\textwidth]{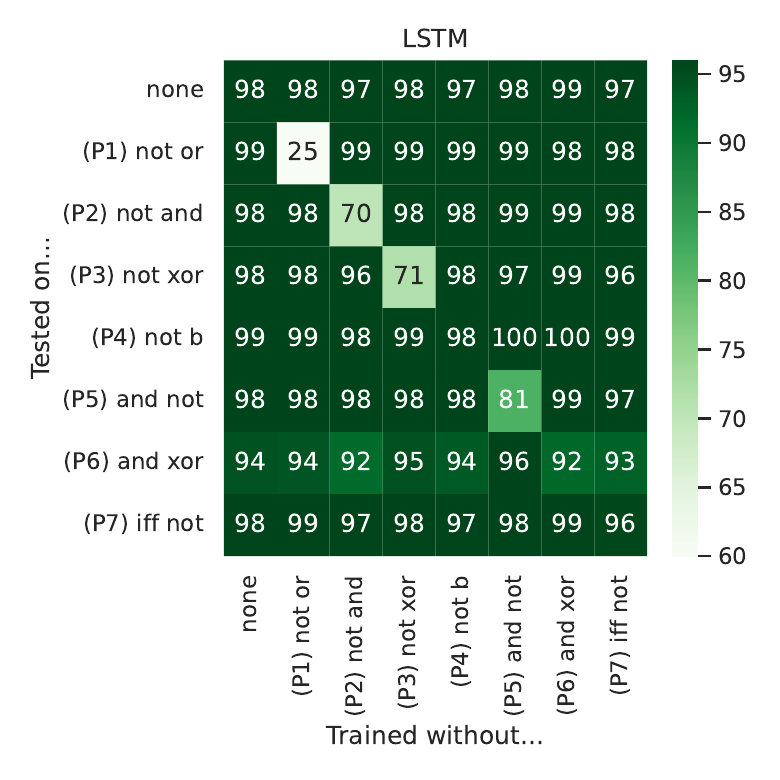}
  \includegraphics[width=.48\textwidth]{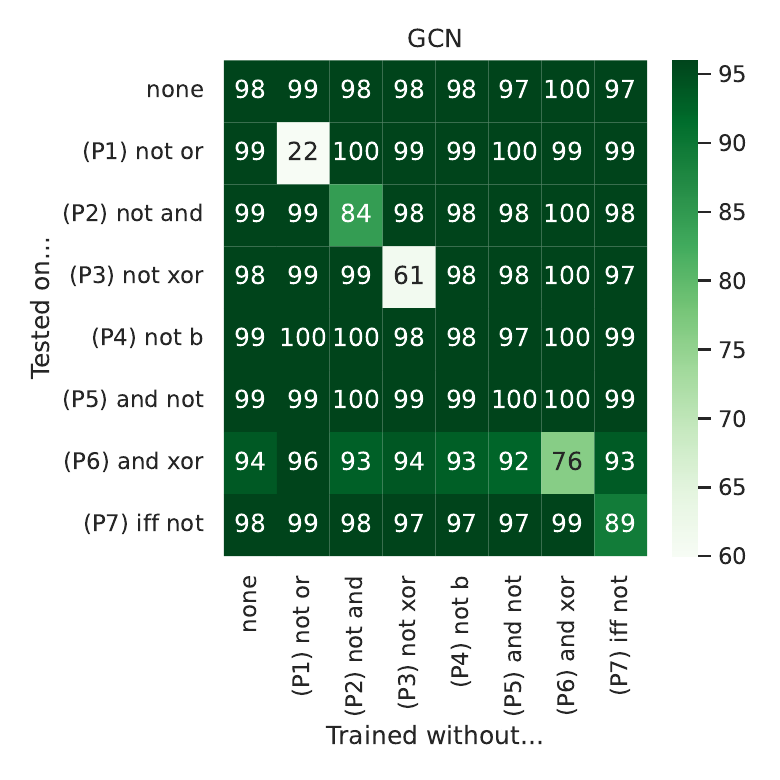}
  \caption{Generalization for all model architectures, on the templated test set. Results are averaged over model seeds. Performance on formulae not containing the held-out pattern is not affected. }
  \label{fig:generalization_detailed}
\end{figure}

\begin{figure}[h]
  \centering
  \includegraphics[width=.8\textwidth]{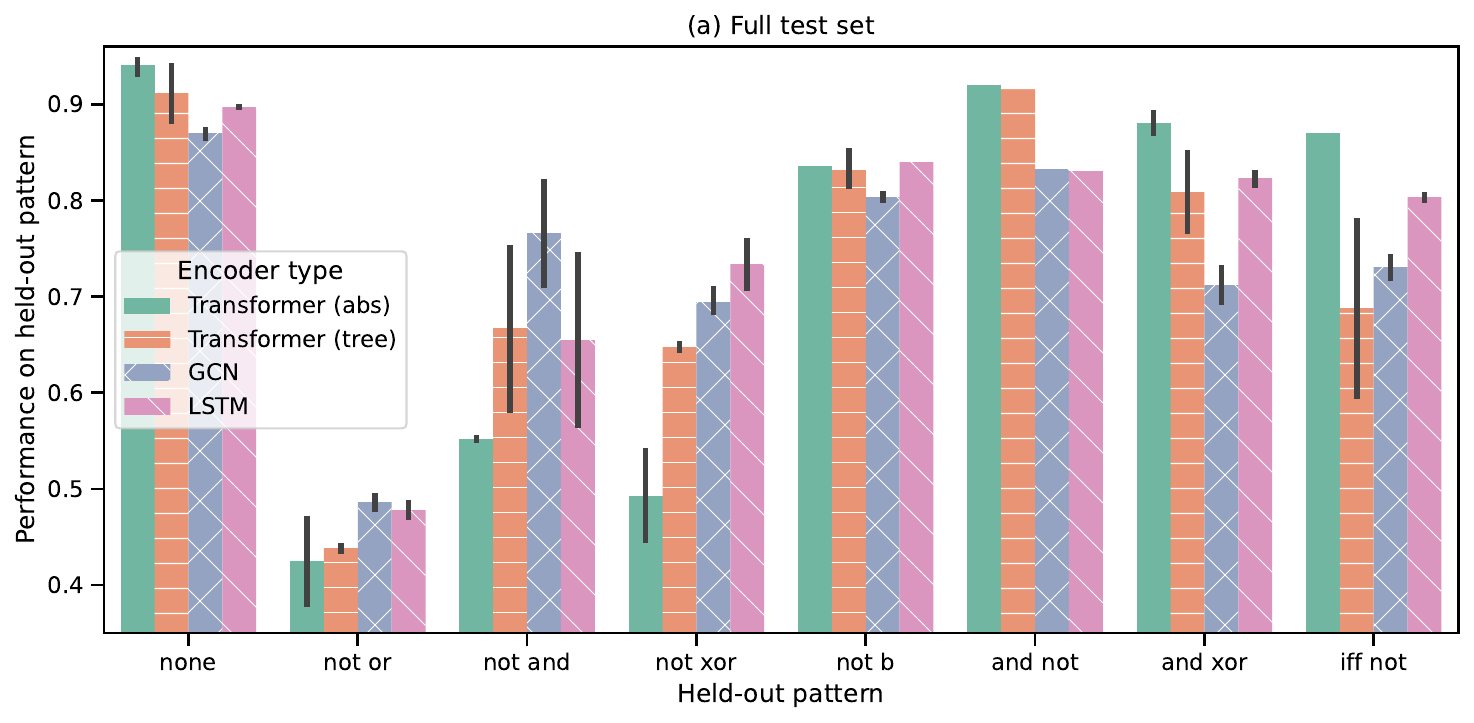}
  \caption{Generalization on the normal test set, for all architectures. Each model (except the base models) is evaluated on sentences containing its respective held-out pattern..}\label{fig:generalization_nontemplate}
\end{figure}

%%% Local Variables:
%%% mode: latex
%%% TeX-master: "main"
%%% End:

\end{document}